\documentclass[cameraready]{Interspeech}

\title{Closing the Speech-Text Gap with Limited Audio for Effective Domain Adaptation in LLM-based ASR}

\author[affiliation={1}, orcid=0009-0003-3247-5908, equalcontribution, correspondingauthor]{Thibault}{Bañeras-Roux}
\author[affiliation={1}, orcid=0000-0002-7694-6834, equalcontribution, correspondingauthor]{Sergio}{Burdisso}
\author[affiliation={1}, orcid=0000-0002-1322-0358, equalcontribution, correspondingauthor]{Esaú}{Villatoro-Tello}
\author[affiliation={1}, orcid=0000-0003-2429-6152]{Dairazalia}{Sánchez-Cortés}
\author[affiliation={1}]{Shiran}{Liu}
\author[affiliation={1, 2}]{Severin}{Baroudi}
\author[affiliation={1,3},orcid=0009-0004-6050-1015]{Shashi}{Kumar}
\author[affiliation={1,3},orcid=0009-0002-2048-8800]{Hasindri}{Watawana}
\author[affiliation={4}]{Manjunath}{K E}
\author[affiliation={4}]{Kadri}{Hacioglu}
\author[affiliation={1,5}, orcid=0000-0001-6467-1119]{Petr}{Motlicek}
\author[affiliation={4}]{Andreas}{Stolcke}

\address{
    $^1$ Idiap Research Institute, Switzerland\\
    $^2$ Laboratoire d'Informatique et des Systèmes, France\\
    $^3$EPFL, Switzerland\\ 
    $^4$ Uniphore, USA \& India\\
    $^5$ Brno University of Technology, Czech Republic
}

\email{\{thibault.roux, sergio.burdisso, esau.villatoro\}@idiap.ch}


\usepackage{comment}

\usepackage{subcaption}
\usepackage{multirow}
\usepackage{xcolor}
\usepackage{colortbl}
\usepackage{tcolorbox}

\usepackage[table]{xcolor}
\definecolor{speech}{HTML}{2E68DB}
\definecolor{keywords}{HTML}{7F00FF}
\definecolor{context}{HTML}{009000}
\definecolor{prompt}{HTML}{666666}
\definecolor{transcript}{HTML}{FFB570}

\newenvironment{promptbox}{
  \begin{tcolorbox}[
    colback=black!1,
    colframe=black!25,
    boxrule=0.4pt,
    arc=2mm,
    left=2mm,right=2mm,top=1mm,bottom=1mm,
  ]
  \ttfamily\footnotesize
  \setlength{\tabcolsep}{0pt}
  
}{
  \end{tcolorbox}
}

\begin{document}

\maketitle

\begin{abstract}
Conventional end-to-end automatic speech recognition (ASR) systems rely on paired speech–text data for domain adaptation. Recent LLM-based ASR architectures connect a speech encoder to a large language model via a projection module, enabling adaptation with text-only data. However, this introduces a modality gap, as the LLM is not exposed to the noisy representations produced by the speech projector. We investigate whether small amounts of speech can mitigate this mismatch. We compare three strategies: text-only adaptation, paired speech–text adaptation, and mixed batching (MB), which combines both. Experiments in in-domain and out-of-domain settings show that even limited speech consistently improves performance. Notably, MB using only 10\% of the target-domain  (less than 4 hours) speech achieves word error rates comparable to, or better than, conventional ASR fine-tuning with the full dataset, indicating that small amounts of speech provide a strong modality-alignment signal.

\end{abstract}

\section{Introduction}
Large language models (LLMs) have transformed speech processing, enabling natural voice interactions. A common approach is LLM-based automatic speech recognition (ASR), where a pretrained speech encoder is connected to a pretrained language model through a lightweight projection layer~\cite{ma2024embarrassingly, wang2023slm, yu2024connecting, yang2025bridging,kumar2025performance,burdisso2025_prompt_projection,villatoro_2026_context_proj}. In particular, SLAM-ASR achieves strong transcription performance without retraining the entire system~\cite{xu2026parameter,zheng2025learning,li2023prompting,baevski2020wav2vec,hsu2021hubert}.

Rather than training from scratch, pretrained systems can be adapted to new domains through fine-tuning on domain-specific data, transferring learned representations~\cite{joshi2022simple, khurana2022magic, huang2020cross}. However, adaptation to specialized domains such as healthcare or finance remains challenging because, although in-domain text is often abundant, labeled speech is expensive to collect.

A standard adaptation strategy is to fine-tune the LLM using in-domain transcripts only~\cite{zheng2025two, fang2025low}. However, this often creates a modality gap~\cite{burdisso2026_TOA,burdisso2026textonlyadaptationllmbasedasr, li2023prompting, fang2025low, ma2024effective}: the model no longer receives representations similar to those produced by the speech projector, weakening speech-text alignment. Existing solutions, such as soft prompting, generally assume a text-only setting~\cite{burdisso2026textonlyadaptationllmbasedasr, fang2025low, ma2024effective} and do not investigate whether limited target-domain speech can help preserve cross-modal alignment.

In this work, we study adaptation across Banking, Insurance, Agriculture, and Musical Instruments domains, analyzing the effect of incorporating limited paired speech-text data through a mixed batching strategy. We compare in-domain and cross-domain adaptation settings to derive practical guidance for scenarios where target-domain text is plentiful, but speech resources are scarce.

Our contributions are fourfold: \textit{(i)} we analyze how the proportion of target-domain data in training batches affects the tradeoff between target-domain specialization and source-domain retention in LLM-based ASR; \textit{(ii)} we propose a hybrid adaptation method that incorporates limited paired speech-text data into a predominantly text-based setting, mitigating the modality gap introduced by text-only fine-tuning; \textit{(iii)} we show that the proposed mixed batching strategy reduces catastrophic forgetting while improving target-domain recognition; and \textit{(iv)} we release our source code for reproducibility.\footnote{\url{https://github.com/idiap/llm-asr-text-only-adaptation/blob/main/README-speech-text-gap.md}}

\section{Related Work}
\label{sec:rel_work}

\textbf{LLM-Based ASR.} %
%
%
Recent work in ASR has shifted toward modular architectures that couple pretrained self-supervised speech encoders such as wav2vec 2.0~\cite{baevski2020wav2vec}, HuBERT~\cite{hsu2021hubert}, or WavLM~\cite{chen2022wavlm}, with LLMs such as LLaMA~\cite{grattafiori2024llama}, or Vicuna~\cite{chiang2023vicuna}), through trainable projection layers that map continuous acoustic features into the LLM embedding space~\cite{ma2024embarrassingly, tangsalmonn, wu2023decoder, ghoshaudio}. 
In this work, we refer to this trainable projection module as the \textit{speech projector}, and to the overall architecture as an \textit{LLM-based ASR} system.
Recent studies demonstrate that such lightweight connectors, ranging from linear projections to Q-Former-style modules, can yield competitive or state-of-the-art WER on benchmarks like LibriSpeech without retraining the entire LLM~\cite{ma2024embarrassingly}. 



\noindent\textbf{Domain Adaptation in Low-Resource Scenarios.}
%
%
Adapting ASR systems to specialized domains such as healthcare, legal, and finance remains a practical necessity, as domain-specific terminology and discourse patterns often diverge significantly from general-purpose training corpora~\cite{zheng2025learning,rangappa2025efficient}. In low-resource scenarios, the primary bottleneck is the scarcity and cost of collecting in-domain speech data~\cite{bhattacharjee2024minimum,ng2026end}, motivating research into text-only domain adaptation. Rather than retraining acoustic models with new audio, these approaches adapt language models or decoding strategies using domain-specific documents such as reports, transcripts, or vocabulary lists. 
More recent work~\cite{li2023prompting, hwang2022large} investigates zero-shot and semi-supervised adaptation, where pretrained ASR models are specialized using textual resources. 
Collectively, these approaches demonstrate that substantial domain-specific WER reductions can be achieved through efficient use of textual data, offering scalable solutions for low-resource deployment.


However, fine-tuning LLMs on clean text degrades their ability to interpret “noisy” representations generated by upstream speech projectors, leading to misalignment and degraded transcription quality~\cite{burdisso2026textonlyadaptationllmbasedasr}. To address this, prior work has explored strategies such as soft prompting~\cite{li2023prompting}, where learned continuous tokens condition the LLM on speech features without rigid tokenization, and soft discretization techniques~\cite{yang2025bridging}. 

\section{Mixed Batching Strategy}
\label{sec:method}


We consider a domain adaptation setting for LLM-based ASR. First, a base model is trained on source-domain paired speech-text data. This base system learns the initial speech–text alignment and serves as the starting point for all experiments. 

We then adapt this pretrained model to a target domain using different fine-tuning strategies. Depending on the setting, the target-domain data may include text-only data, paired speech-text data, or a combination of both. The goal of adaptation is to improve recognition performance on the target domain while preserving performance on the original source domain.

To mitigate the modality gap introduced by text-only adaptation, we build upon the mixed batch composition strategy proposed
in a previous study~\cite{burdisso2026textonlyadaptationllmbasedasr}.
In that work, the authors integrate target-domain text-only data with source-domain auxiliary data during fine-tuning to preserve speech-text alignment and mitigate catastrophic forgetting while enabling adaptation without target-domain audio, by dividing each batch into a source portion (auxiliary data) and a target-domain portion

In our experiments, we extend this strategy by integrating target-domain audio into the batch composition. 
More precisely, similar to~\cite{burdisso2026textonlyadaptationllmbasedasr}, the source data in the batch is composed of three segments:

(1) \textbf{Paired source speech-text} (batch proportion $\sigma_a$): the input to the model is a source-domain audio utterance, and the output is its corresponding transcript. This component allows the model to preserve the original audio-text alignment learned during base training.

(2)~\textbf{Projector-induced noisy tokens} (batch proportion $\sigma_{ta}$): the input to the model is a discrete sequence obtained by mapping each speech projection to its nearest tokens in the LLM embedding space, and the output is the corresponding clean transcript. This helps the model to understand discrete text from projected speech representations.

(3)~\textbf{Synthetically corrupted source transcripts} (batch proportion $\sigma_t$): the input to the model is a corrupted transcript generated through random character-level perturbations using \texttt{nlpaug}\footnote{\url{https://nlpaug.readthedocs.io/en/latest/augmenter/char/random.html}} (with the same parameters as in the original paper), and the output is the original clean transcript. This discrete input sequence simulates the irregularities observed in projected speech-token representations without requiring audio.


Following the paired speech-text and synthetically corrupted transcripts, the target portion of the batch consists of:

(1)~\textbf{Paired target speech-text} (batch proportion $\tau_a$): the input to the model is a target-domain audio utterance, and the output is its corresponding transcript. This component induces the adaptation in real target-domain acoustics and preserves speech-text alignment.

(2)~\textbf{Corrupted target transcripts} (batch proportion $\tau_t$): the input to the model is a corrupted target-domain transcript generated through the same character-level perturbations as in $\sigma_t$, and the output is the corresponding clean transcript. This component supports domain-specific adaptation using text-only supervision.
    

While $\tau_t$ controls domain-specific language modeling, $\tau_a$ keeps the adaptation grounded in real target-domain speech and acoustics. From now on, we will refer the strategy described here as Mixed Batch (MB).


\section{Method}
\label{sec:protocol}



\subsection{LLM-based ASR Architecture}
\label{sec:asr-architecture}

In our experiments, the ASR system follows the SLAM-ASR framework~\cite{ma2024embarrassingly}. The architecture consists of three main components: (i) a pretrained speech encoder, (ii) a trainable speech projector, and (iii) an LLM.

Given a speech segment $X^S$, the speech encoder produces a sequence of acoustic representations $H^S = [h_1^S, \dots, h_T^S]$ over $T$ time frames. To better match the token-level processing of the LLM, this sequence is temporally downsampled by a factor $k$, resulting in $Z^S = [z_1^S, \dots, z_N^S]$ with $N = T // k$. 

The downsampled representations are then passed through a linear speech projector, which maps them into the LLM embedding space, producing $E^S$. These projected embeddings are directly consumed by the LLM, together with a task-specific prompt that conditions the model for speech recognition. The LLM then autoregressively generates the output transcript.

In our setup, we first train a base model using paired speech-text data from the source corpus (see Section~\ref{sec:datasets}). During this stage, the speech encoder is kept frozen, while the speech projector is fine-tuned to learn the alignment between acoustic representations and the LLM embedding space. This results in a base ASR system adapted to the source domain.

The base model is subsequently adapted to target domains using the various strategies described, by fine-tuning only LoRA adapters~\cite{hu2022lora}. During adaptation, the speech encoder and the speech projector are frozen, and only the LoRA modules inserted into the LLM are updated. This design enables parameter-efficient specialization to the target domain while preserving the previously learned speech–text alignment.

For both base training and adaptation, models are trained for 5 epochs with a batch size of 10. The speech encoder is \texttt{WavLM-Large}~\cite{chen2022wavlm}, and the LLM is \href{https://huggingface.co/meta-llama/Llama-3.2-3B-Instruct}{\texttt{Llama-3.2-3B-Instruct}}~\cite{grattafiori2024llama}.

All experiments use a fixed instruction-style prompt. Specifically, the LLM input follows the template:
\begin{promptbox}
\ttfamily\footnotesize
\textbf{\textcolor{prompt}{<s>USER: Transcribe speech to text.}\\
\textcolor{prompt}{Speech:}\textcolor{speech}{\{speech\}}\\
\textcolor{prompt}{ASSISTANT:} \itshape{\textcolor{transcript}{\{transcript\}}}\textcolor{prompt}{</s>}}
\end{promptbox}
\noindent
where \textcolor{speech}{\texttt{\textbf{\{speech\}}}} corresponds to the projected speech embeddings. This prompt conditions the LLM to perform transcription in an instruction-following setup consistent with its pretraining.

\begin{table}[t]
\setlength{\tabcolsep}{1.8pt}
\rowcolors{3}{gray!10}{white} 
\caption{Datasets used for training, validation, and testing in our experiments. From the DefinedAI corpus, we use Banking (B), Insurance (I) and Healthcare (H) partitions. From the SlideSpeech corpus, we use Life (L), Talent (T), English (E), Agriculture (Ag), Animation (An) and Musical Instruments (MI). We provide the total number of utterances (\#Utts) and the duration (Hrs) of each partition.}
\begin{tabular}{lccccccc}
\hline
\multirow{2}{*}{\textbf{Dataset}} & \multirow{2}{*}{\textbf{Domain}} & \multicolumn{2}{c}{\textbf{Train}} & \multicolumn{2}{c}{\textbf{Dev}} & \multicolumn{2}{c}{\textbf{Test}} \\
 &  & \textbf{\#Utts} & \textbf{Hrs} & \textbf{\#Utts} & \textbf{Hrs} & \textbf{\#Utts} & \textbf{Hrs} \\ \hline
\textit{\textbf{Source}} &  &  &  &  &  &  &  \\
DefinedAI & B/I/H & 17,398 & 38h10 & -- & -- & -- & -- \\
Slidespeech & L/T/E & 34,692 & 34h59 & -- & -- & -- & -- \\
\textit{\textbf{Target}} &  &  &  &  &  &  &  \\
DefinedAI & B & 26,704 & 36h20 & 1,625 & 2h08 & 3,111 & 4h16 \\
DefinedAI & I & 32,249 & 46h54 & 1,951 & 2h45 & 3,807 & 5h30 \\
SlideSpeech & Ag & 30,498 & 29h20 & 1,679 & 1h43 & 3,470 & 3h26\\
SlideSpeech & An & 56,593 & 49h48 & 3,283 & 2h55 & 7,005 & 5h50\\
SlideSpeech & MI & 9,981 & 8h30 & 852 & 0h44 & 984 & 0h54 \\ \hline
\end{tabular}
\label{tab:datasets}
\end{table}

\subsection{Datasets and Domain Configuration}
\label{sec:datasets}


Our experiments leverage two corpora, DefinedAI\footnote{\url{https://defined.ai/}} and SlideSpeech~\cite{wang2024slidespeech}.  
The source-domain data comes from DefinedAI, encompassing Banking (B), Insurance (I), and Healthcare (H) subsets. This dataset is used to pretrain the projector and serve as auxiliary data during LLM adaptation.

The target-domain data consists of specific subsets from both corpora. From DefinedAI, we select the Banking partition to train and evaluate in-domain adaptation. SlideSpeech contributes data from the Agriculture (Ag) and Musical Instruments (MI) domains, representing out-of-distribution scenarios with limited labeled audio. Table~\ref{tab:datasets} summarizes the statistics for each dataset, including the number of utterances and total audio duration for training, development, and test splits.

\section{Results}

Before conducting the adaptation experiments, we first train a base ASR model exclusively on the source-domain data from the DefinedAI corpus, including the Banking, Insurance, and Healthcare partitions. This base model serves as the starting point for all subsequent experiments. We then perform domain adaptation on different target domains: (i) the in-domain DefinedAI Banking set, and (ii) out-of-distribution domains from the SlideSpeech corpus, namely Agriculture (Ag) and Musical Instruments (MI). All adaptation strategies described in the following sections are applied to this same base model to ensure controlled comparisons across domains and training settings.

\subsection{Target Data Proportion in Text-Only Adaptation}
\label{sec:exp-tau-t}


We investigate the effect of varying the proportion of target text data per batch ($\tau_t$) in the text-only adaptation setting. No audio is included ($\tau_a = 0$); the remaining portion of a batch is split uniformly among the various types of source-domain text data.

\begin{figure}[t!]
  \centering
  \includegraphics[width=\linewidth]{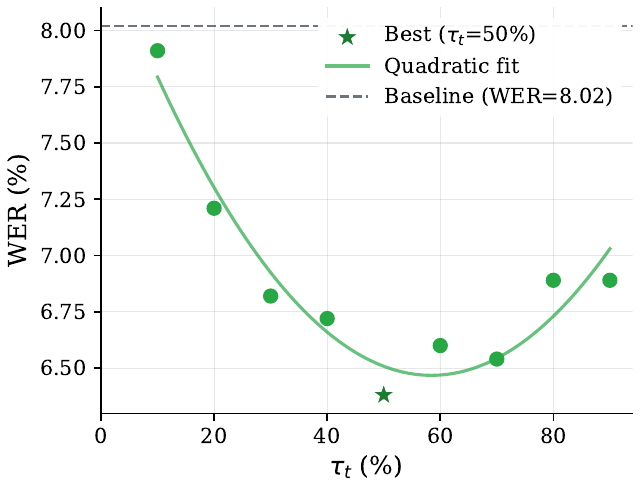}
  \caption{Effect of different target text proportions ($\tau_t$) on mixed-batch adaptation on DefinedAI Banking.}
  \label{fig:tau-t-impact}
\end{figure}

\begin{figure*}[!t]
  \centering
  \begin{subfigure}[b]{0.32\linewidth}
    \centering
    \includegraphics[width=\linewidth]{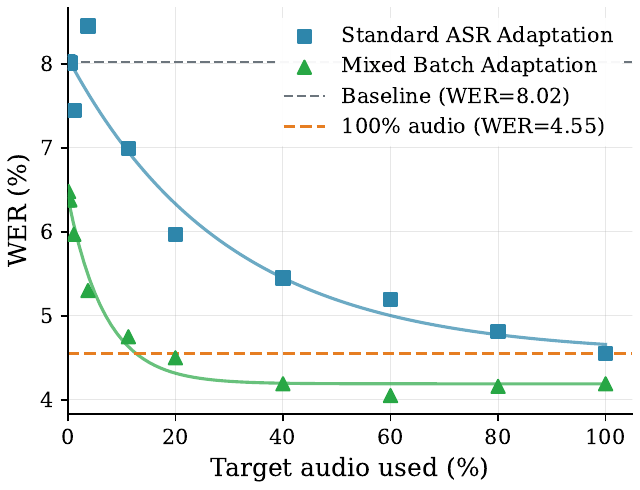}
    \caption{Banking}
    \label{fig:audio-vs-ta-banking}
  \end{subfigure}
  \hfill
  \begin{subfigure}[b]{0.32\linewidth}
    \centering
    \includegraphics[width=\linewidth]{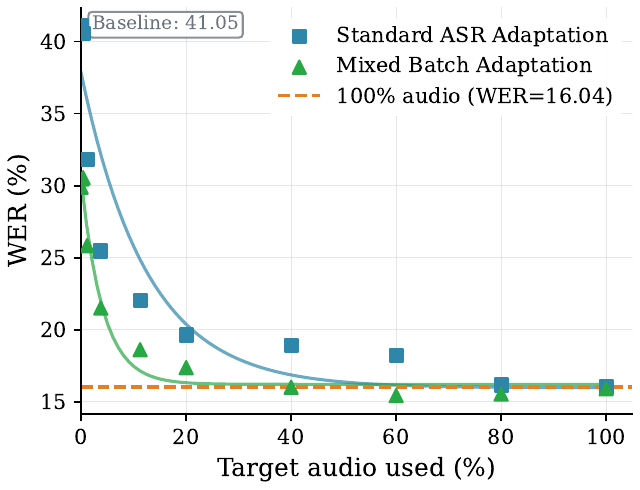}
    \caption{Agriculture}
    \label{fig:audio-vs-ta-ag}
  \end{subfigure}
  \hfill
  \begin{subfigure}[b]{0.32\linewidth}
    \centering
    \includegraphics[width=\linewidth]{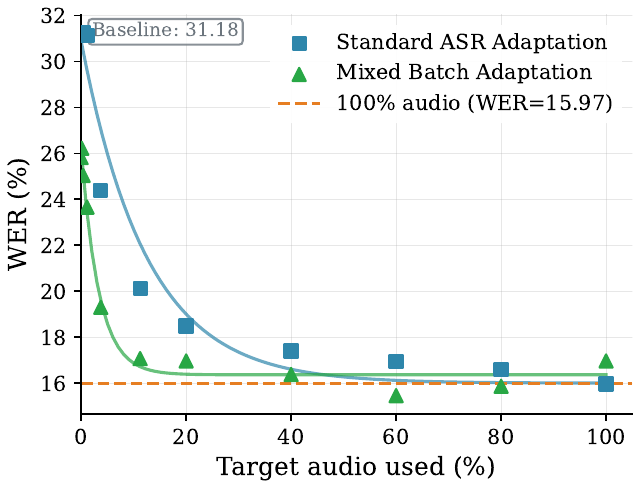}
    \caption{Musical Instruments}
    \label{fig:audio-vs-ta-mi}
  \end{subfigure}
  \caption{Performance of the base ASR and adapted models. The figure compares standard ASR adaptation using only paired speech–text data with mixed-batch adaptation combining paired speech–text and text-only samples. The x-axis indicates the proportion of speech in the adaptation data, ranging from 0\% (text-only) to 100\% (paired speech–text).}
\label{fig:audio-vs-ta}
\end{figure*}

As shown in Figure~\ref{fig:tau-t-impact}, performance follows a clear tradeoff pattern. Increasing $\tau_t$ initially improves recognition accuracy, reaching optimal performance around $\tau_t = 50\%$. Beyond this point, performance degrades, reflecting a balance between: (1) Low $\tau_t$: excessive reliance on auxiliary data, which may be repeatedly sampled and lead to overfitting. (2) High $\tau_t$: insufficient exposure to auxiliary data, potentially causing underfitting and partial catastrophic forgetting.

Based on these results, we adopted $\tau = \tau_t + \tau_a = 50\%$ in all experiments and $\sigma$ values being uniformly distributed, as that seems to provide the best performance.


\subsection{Paired Speech-Text vs.\ Mixed-Batch Adaptation}
\label{sec:exp-audio-vs-ta-audio}

We now look at whether text-only adaptation combined with a small amount of audio can reduce the modality gap between text-only and paired speech-text training.
%
Both in previous studies~\cite{burdisso2026textonlyadaptationllmbasedasr,xu2026parameter, fang2025low} and in our current setup
, we observe an important performance difference between the text-only adapted model and the speech-text adapted model. In particular, in our experience, paired speech-text adaptation outperforms text-only adaptation by 1.83\% in absolute WER ($\approx$ 29\% relative improvement) on the Banking domain (4.55\% vs. 6.38\%), highlighting a substantial modality gap.
We therefore investigate whether incorporating a small quantity of target-domain audio during adaptation can effectively bridge this gap, and if the system can get close to the best-case scenario where we have full audio.
More specifically, we compare:
\begin{itemize}
    \item \textbf{Speech-text adaptation}, or \textit{standard ASR adaptation}, where the model is fine-tuned exclusively on various amount of labeled speech with the associated transcript.
    \item \textbf{Mixed-batch adaptation}, where text-based adaptation is complemented with a small quantity of audio speech with associated transcript (\textit{i.e.}, text-only and paired speech-text together comprise the full training set).
\end{itemize}


We fine-tune the base model using different fractions of each corpus, including setups with very limited speech data. This allows us to study how the model adapts under low-resource conditions, where the available audio alone is insufficient to achieve strong performance, as well as under scenarios with progressively more adaptation speech. 

Figure~\ref{fig:audio-vs-ta} shows that 
mixed-batch adaptation consistently outperforms paired speech-text training. Adding speech to a text-adapted model leads to larger gains than training on the same amount of paired speech-text. With only 10\% of the adaptation corpus containing speech (corresponding to 3h37, 2h55, and 50m for Banking, Agriculture, and Musical Instruments, respectively), 
we obtain performance comparable or better than using the full speech corpus. This suggests that text adaptation can improve an LLM's domain-specific abilities, while additional speech helps align audio representations with the embedding space of the LLM, thereby reducing the modality gap.

Across all target domains, we observe that mixed-batch adaptation achieves its peak performance at $\sim$60\% of the corpus consists of paired speech-text examples, after which the gains slightly decrease. We hypothesize that this behavior arises from the presence of synthetically generated noisy transcripts in the batch, which introduce additional diversity in the inputs, seemingly helping the model to learn more robust mappings between speech and text representations. Mixed batching seems not only leverages the benefits of target-domain text adaptation, but could improve generalization under low-resource conditions.

However, as the amount of available speech increases, the performance difference between the two strategies gradually diminishes. When sufficient speech data is available, both approaches converge toward similar recognition accuracy (with respect to mixed-batch adaptation and paired speech-text adaptation: \textbf{4.19\%} vs.\ \textbf{4.55\%} for Banking, \textbf{15.91\%} vs.\ \textbf{16.04\%} for Agriculture, \textbf{16.97\%} vs.\ \textbf{15.97\%} for Musical Instruments), indicating that audio supervision alone eventually becomes sufficient. 
For in-domain adaptation (Banking), mixed-batch adaptation still shows a slight advantage, likely due to the contribution of auxiliary data drawn from the same domain.

\subsection{Analysis of Catastrophic Forgetting}
\label{sec:exp-castrophic-forgetting}

To assess the impact of adaptation on previously learned knowledge, we evaluate all systems on both source-domain and target-domain test sets. In this analysis, the source domain corresponds to the DefinedAI (B/I/H) data used for base training, while the target domain is the DefinedAI Banking subset. This setup allows us to quantify potential catastrophic forgetting.

\begin{figure}[h]
  \centering
  \includegraphics[width=\linewidth]{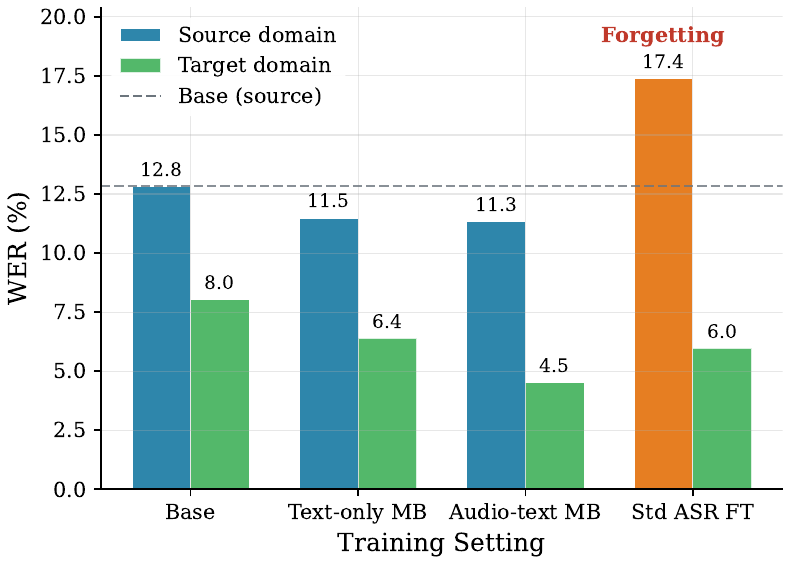}
  \caption{Performances of base system and adapted model with text-only, paired speech-text and mixed-batch adaptation. For both audio-adaptation settings, the share of speech data is 20\%.}
  \label{fig:catastrophic_forgetting}
\end{figure}

As shown in Figure~\ref{fig:catastrophic_forgetting}, the base model achieves a WER of 12.8\% on the source domain and 8.0\% on the target domain. 
This indicates that, in our setup, the target test set is easier than the source data for speech recognition. When adapting using paired speech-text (with 20\% of target audio), we observe a degradation in the source domain, even though the adaptation remains in-domain. While recognition improves on the target domain, the accuracy drop in the source domain is evidence of catastrophic forgetting. In contrast, text-only and mixed-batch adaptation result in an improvement in target-domain performance while limiting degradation on the source domain.

\section{Conclusions}

In this work, we have investigated domain adaptation for LLM-based ASR in settings where target-domain speech is limited, while textual data is abundant. We identified the modality mismatch introduced by text-only fine-tuning as a key factor behind performance degradation and proposed a hybrid batching strategy that mixes speech-text pairs with text to address this issue.

By integrating auxiliary source data with both corrupted target text and a controlled proportion of target-domain audio, our approach preserves speech-text alignment while enabling effective domain adaptation. Through systematic experiments, we showed that using our hybrid batch method, even with modest amounts of target-domain audio, substantially improves recognition accuracy and mitigates catastrophic forgetting. Notably, combining text data with a small fraction of audio can outperform adaptation using the full audio dataset, highlighting the data efficiency of the proposed strategy.

We also observed a clear tradeoff between adapting to the target domain and preserving performance on the source domain, with respect to batch composition. Our findings provide practical modeling guidelines for real application scenarios. 

\section{Acknowledgments}

This work was supported by an Idiap Research Institute and Uniphore collaboration project. Part of this work was also supported by EU Horizon 2020 project ELOQUENCE.

\section{Generative AI Use Disclosure}
Generative AI was used to proofread the paper, fix orthographic and grammatical errors, and shorten the length of some sections.

\section{Appendices}

\begin{figure*}[!h]
  \centering
  \begin{subfigure}[b]{0.32\linewidth}
    \centering
    \includegraphics[width=\linewidth]{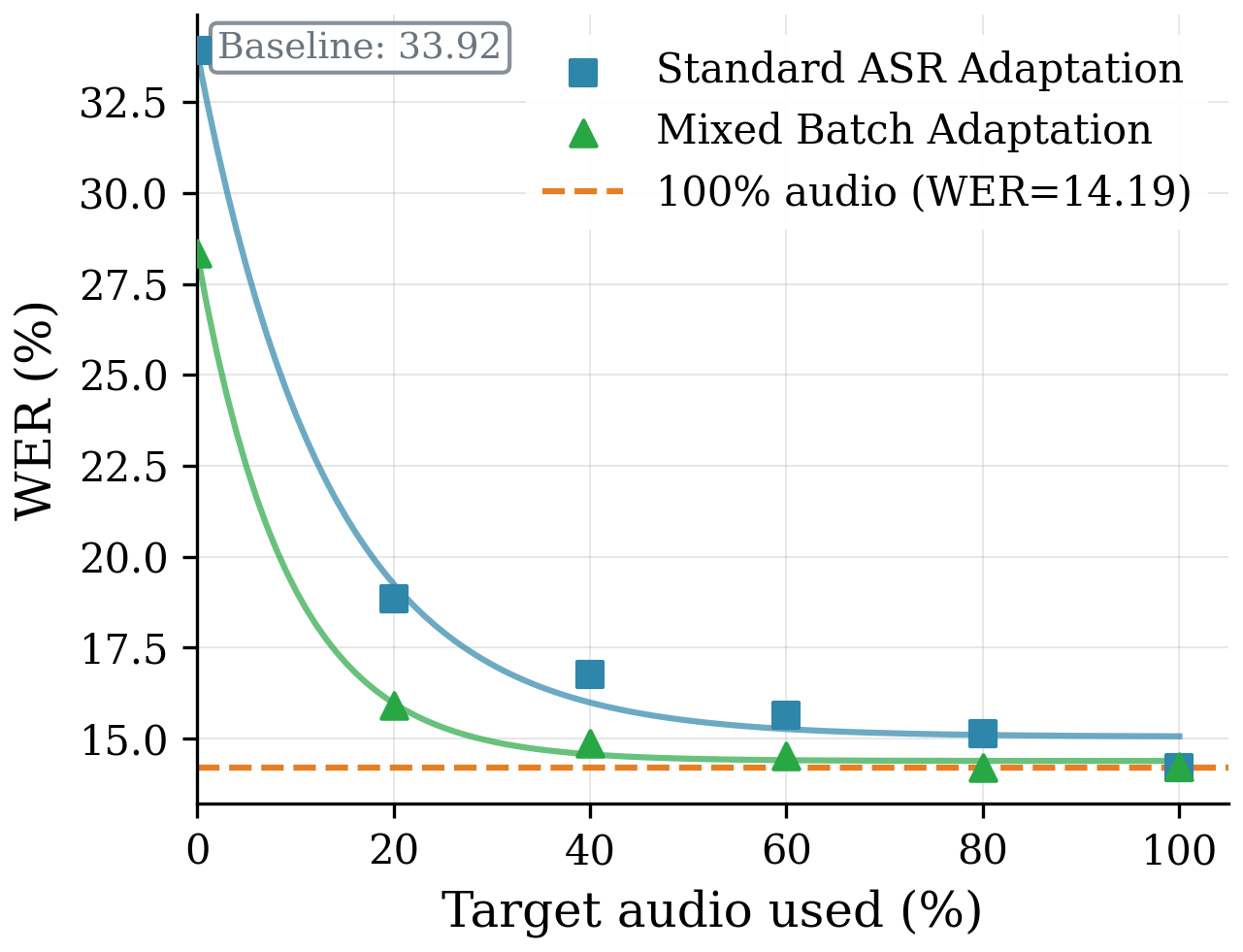}
    \caption{Animation}
    \label{fig:audio-vs-ta-dai-an}
  \end{subfigure}
  \begin{subfigure}[b]{0.32\linewidth}
    \centering
    \includegraphics[width=\linewidth]{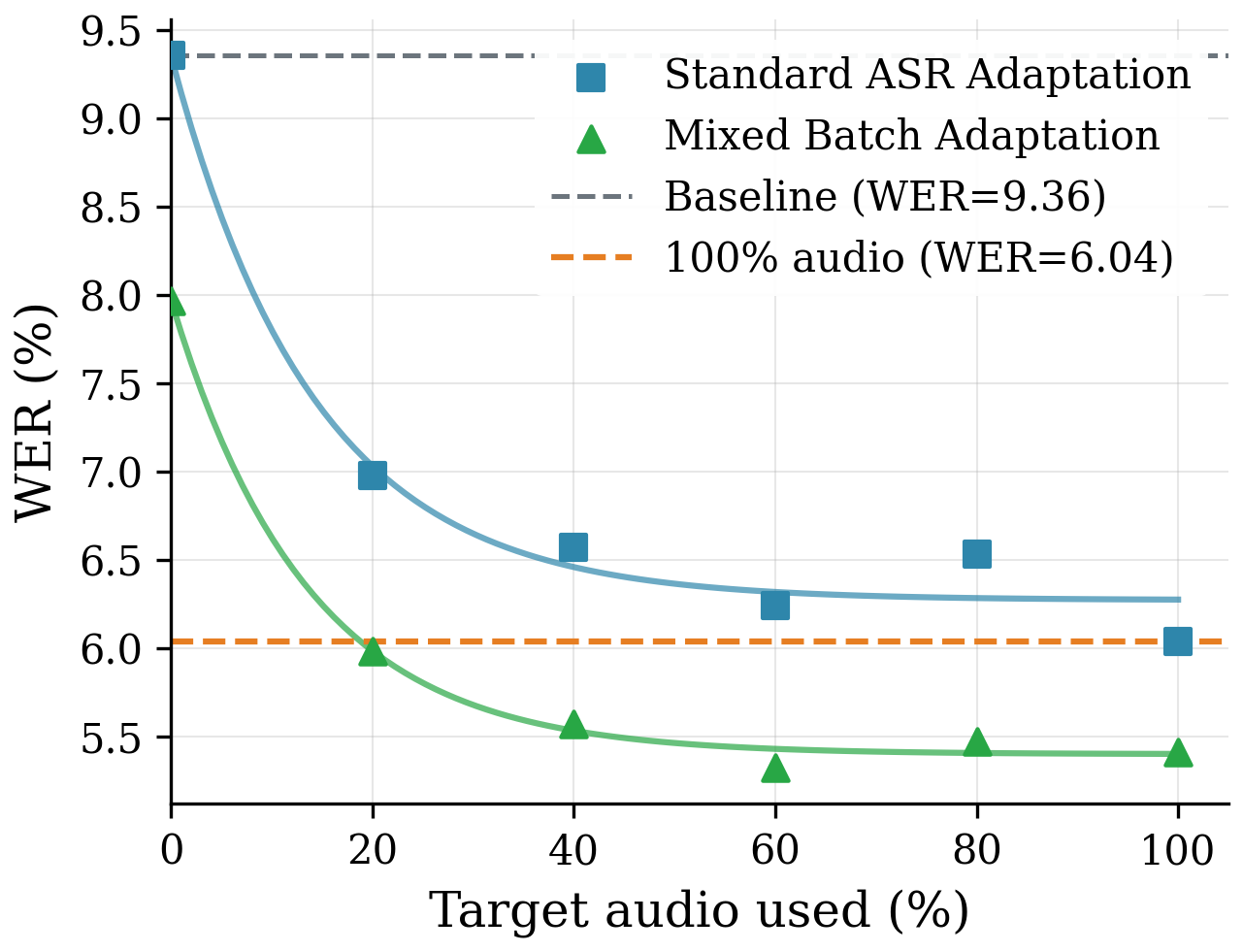}
    \caption{Insurance}
    \label{fig:audio-vs-ta-dai-i}
  \end{subfigure}
  \caption{Performance of the base ASR pretrained on DefinedAI and adapted models. The figure compares standard ASR adaptation using only paired speech–text data with mixed-batch adaptation combining paired speech–text and text-only samples. The x-axis indicates the proportion of speech in the adaptation data, ranging from 0\% (text-only) to 100\% (paired speech–text).}
\label{fig:audio-vs-ta-complementary}
\end{figure*}
\begin{figure*}[!h]
  \centering
  \begin{subfigure}[b]{0.32\linewidth}
    \centering
    \includegraphics[width=\linewidth]{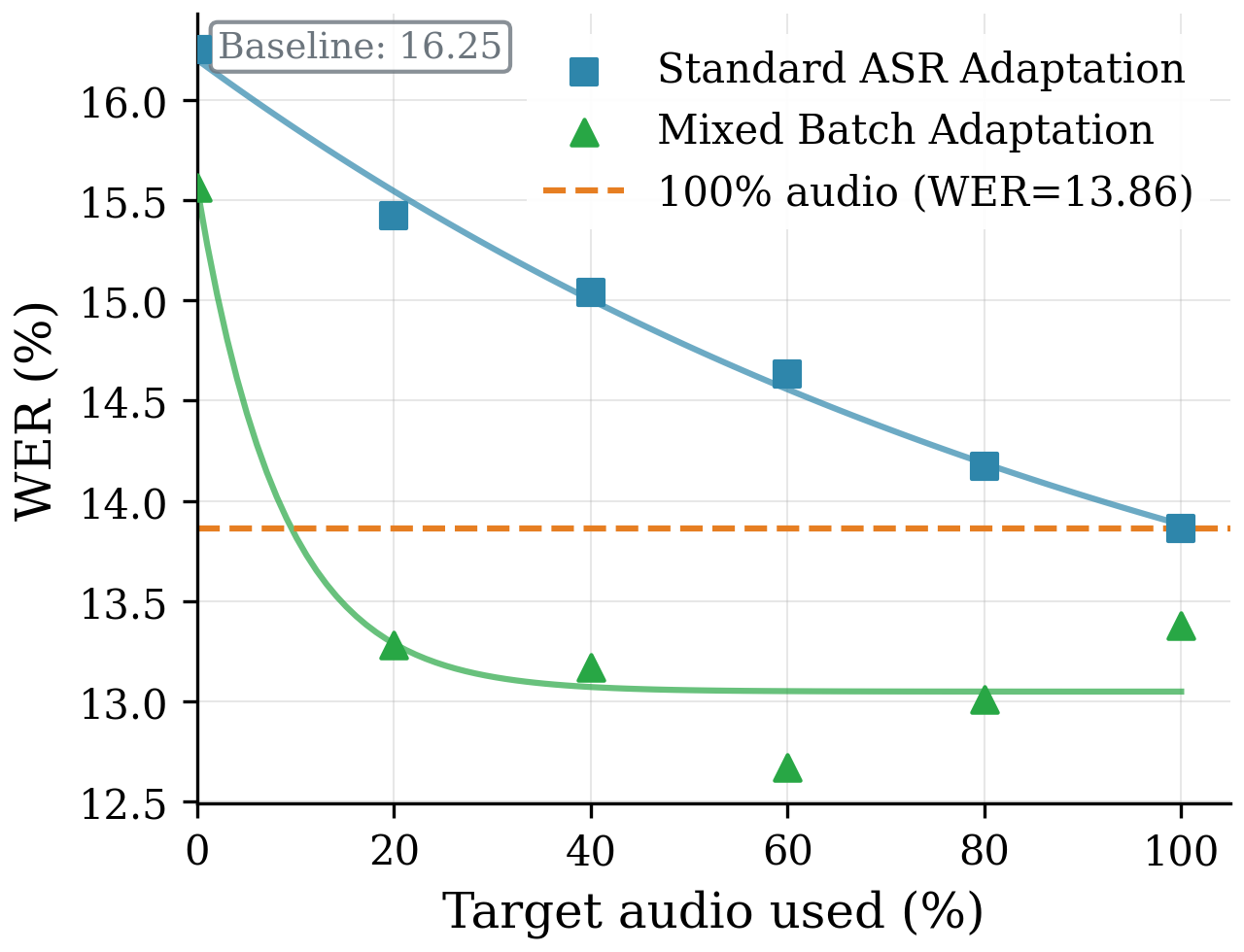}
    \caption{Agriculture}
    \label{fig:audio-vs-ta-ss-ag}
  \end{subfigure}
  \hfill
  \begin{subfigure}[b]{0.32\linewidth}
    \centering
    \includegraphics[width=\linewidth]{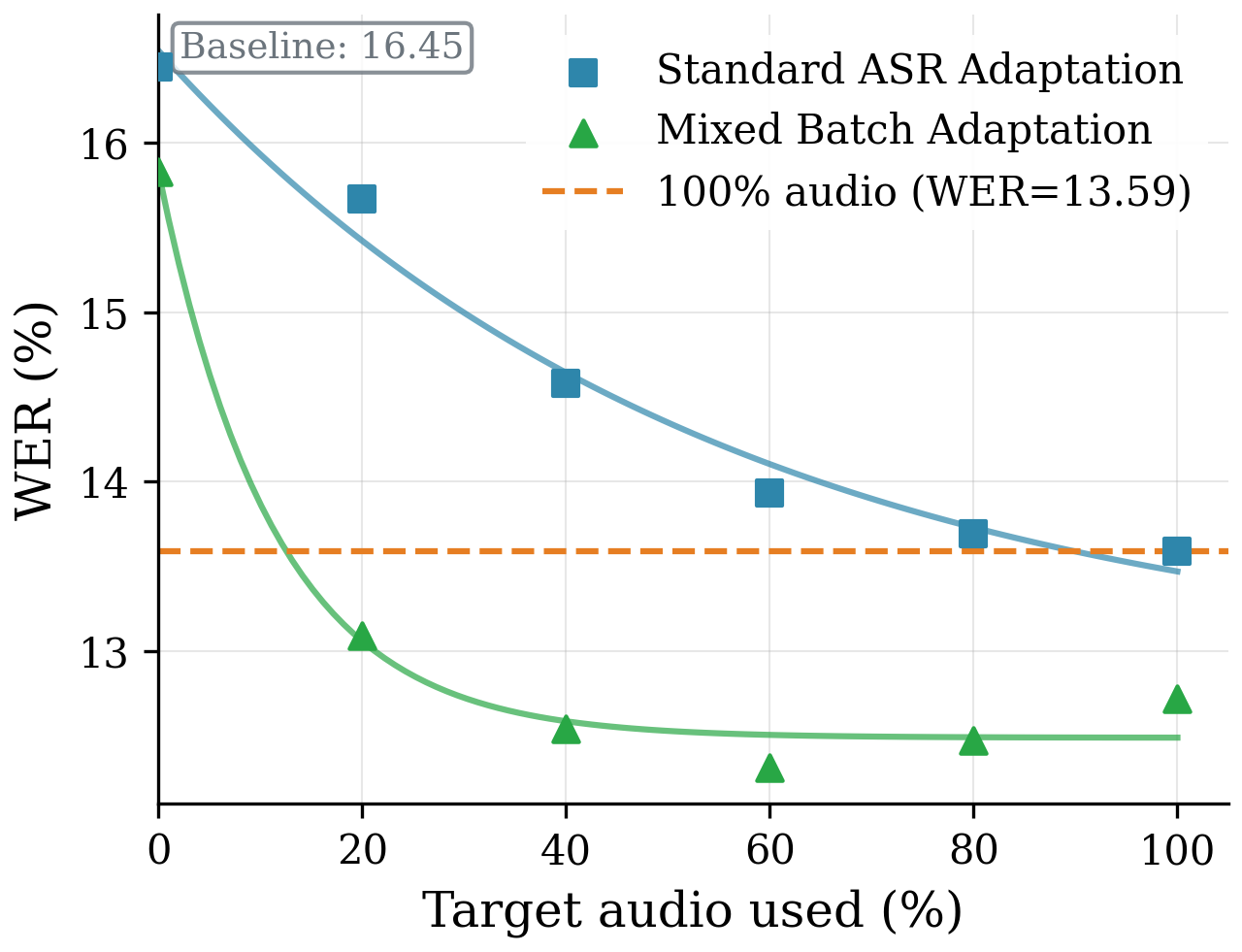}
    \caption{Animation}
    \label{fig:audio-vs-ta-ss-an}
  \end{subfigure}
  \hfill
  \begin{subfigure}[b]{0.32\linewidth}
    \centering
    \includegraphics[width=\linewidth]{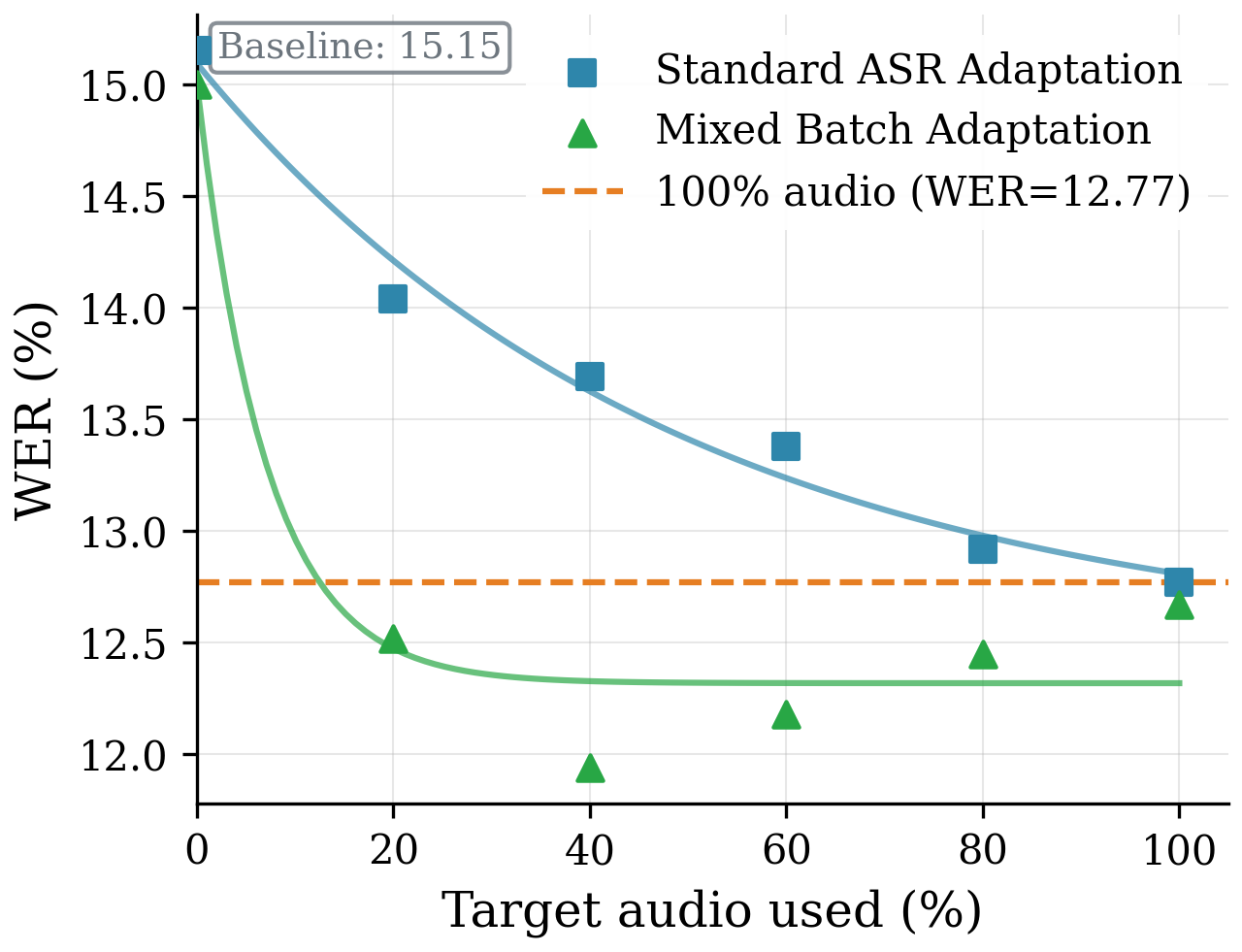}
    \caption{Musical Instruments}
    \label{fig:audio-vs-ta-ss-mi}
  \end{subfigure}
  \caption{Performance of the base ASR pretrained on Slidespeech and adapted models. The figure compares standard ASR adaptation using only paired speech–text data with mixed-batch adaptation combining paired speech–text and text-only samples. The x-axis indicates the proportion of speech in the adaptation data, ranging from 0\% (text-only) to 100\% (paired speech–text).}
\label{fig:audio-vs-ta-complement-ss}
\end{figure*}

To further assess the trends observed in the main experiments, Figures~\ref{fig:audio-vs-ta-complementary} and~\ref{fig:audio-vs-ta-complement-ss} report the same comparison on additional domains and datasets. The results consistently support the findings discussed in Section~\ref{sec:exp-audio-vs-ta-audio}: mixed-batch adaptation provides substantial gains when only a small amount of paired speech-text data is available, often reaching performance comparable to that obtained with substantially more speech. As the proportion of paired speech-text data increases, the performance gap between the two adaptation strategies gradually narrows. These additional results therefore reinforce the observation that combining domain-specific text with a limited amount of speech can effectively reduce the modality gap and improve adaptation in low-resource settings.

\bibliographystyle{IEEEtran}
\bibliography{mybib}

\end{document}